%% file: CameraReady2027.tex
\documentclass[letterpaper]{article} 
\usepackage{aaai2027}  
\usepackage[hyphens]{url}  
\usepackage{graphicx} 
\usepackage{natbib}  
\usepackage{caption} 
\usepackage{algorithm}
\usepackage{algorithmic}
\usepackage{multirow}
\usepackage{amsmath}
\usepackage{amssymb}

\usepackage{newfloat}
\usepackage{listings}
\DeclareCaptionStyle{ruled}{labelfont=normalfont,labelsep=colon,strut=off} 
\floatstyle{ruled}
\newfloat{listing}{tb}{lst}{}
\floatname{listing}{Listing}

\usepackage{booktabs}

\title{S³martCirc: Self-supervised Smart Circuit Discovery}
\author {
    Wendy Zheng\textsuperscript{\rm 1}\equalcontrib,
    Yinhan He\textsuperscript{\rm 1}\equalcontrib,
    Liang Wu\textsuperscript{\rm 2}
    Jundong Li\textsuperscript{\rm 1}\corresponding
}
\affiliations {
    \textsuperscript{\rm 1}University of Virginia\\
    \textsuperscript{\rm 2}Nokia\\
    ncd9cf@virginia.edu,nee7ne@virginia.edu, liang.wu@nokia.com, jundong@virginia.edu
}

\begin{document}

\maketitle

\begin{abstract}
\input{sections/0-abstract}
\end{abstract}

\begin{links}
     \link{Code}{https://anonymous.4open.science/r/s3martcirc}
\end{links}

\section{Introduction}
\input{sections/1-intro}

\section{Preliminaries and Definitions}\label{sec:preliminaries}
\input{sections/2-preliminaries}

\section{Self-supervised Smart Circuit Discovery}
\input{sections/3-method}

\section{Experiments}
\input{sections/4-experiments}

\section{Related Works}
\input{sections/5-related_works}

\section{Conclusion}
\input{sections/6-conclusion}

\bibliography{aaai2027}

\clearpage
\newpage
\appendix
\input{sections/appendix}


\end{document}

%% file: sections/0-abstract.tex
Large Language Models (LLMs) have demonstrated remarkable performance across diverse tasks, from text summarization to question answering. Despite these capabilities, their black-box nature obscures internal decision-making processes. Mechanistic interpretability (MI) aims to address this by reverse-engineering neural networks into human-understandable algorithms. Current MI approaches for LLMs typically follow a two-stage paradigm: first identifying important components (circuit discovery), where components are typically individual nodes such as an attention head or feedforward neuron, and second determining the role they play in a certain task (functional interpretation). However, this sequential approach overlooks a fundamental insight: a component's importance and its functional role are inherently codependent. Unifying these stages presents two key challenges: (1) functional roles are often tied to specific nodes or components, limiting generalization, and (2) their identification relies on subjective interpretation rather than quantifiable metrics. To address these challenges, we propose S³martCirc (\textbf{S}elf-\textbf{s}upervised \textbf{S}mart \textbf{Cir}cuit Discovery), a unified framework that simultaneously discovers circuits and interprets functionality. S³martCirc abstracts node behavior into two general computational roles that generalize across tasks and defines a quantitative metric for assigning them, enabling importance and functional role to be discovered jointly rather than in sequence. Extensive experiments show that our framework outperforms existing methods in circuit discovery.

%% file: sections/1-intro.tex
Recently, Large Language Models (LLMs) have exhibited extraordinary performance across a wide range of diverse tasks, from text summarization~\cite{ basyal2023text,van2023clinical} and translation~\cite{feng2024tear,kleidermacher2025science} to question answering~\cite{li2024flexkbqa,tan2023can,yang2023one}. However, these models are primarily black boxes whose internal reasoning mechanisms cannot be directly accessed or interpreted by humans. This opacity obscures their decision-making processes and raises safety concerns about potential undesired behaviors~\cite{singh2024rethinking, wagner2024black}. Consequently, it hinders the adoption of LLMs in critical fields such as healthcare and finance, where reliability and trustworthiness are paramount~\cite{tatsat2025beyond,yang2023large}.

To better understand how LLMs operate, various interpretation techniques have been proposed~\cite{chenprobing,he2025saif,hoover2020exbert,singh2024rethinking}, among which mechanistic interpretability (MI) has emerged as a particularly promising direction~\cite{bereska2024mechanistic, sharkey2025open}. MI aims to reverse engineer neural networks into human-understandable algorithms by uncovering how components collaborate to implement a specific task. Current approaches for LLMs generally follow a two-stage paradigm: (1) identify important nodes (i.e., attention heads or feedforward neurons) through circuit discovery, and (2) determine the functional role each selected node performs through functional interpretation~\cite{liu2025large,nanda2023progress,wang2022interpretability}. 

An alternative perspective suggests that this sequential separation is not fundamental. The order of the two stages can be reversed: one may first identify the functional operations required for a task and subsequently search for the nodes that implement them~\cite{melouxeverything}. In this view, node importance and functionality are intrinsically interdependent: a node is important because of the function it performs, while its function is defined by its contribution to the task. Knowing a node's functional role therefore provides direct evidence for whether it is relevant to the task, and conversely, knowing that a node is important constrains the role it is likely to play. This mutual dependence implies that circuit discovery and functional interpretation should not be treated as sequential steps, but as a single, jointly optimized problem. However, realizing this perspective faces two key challenges: (1) \textbf{Node-specific interpretations.} Prior works assign fine-grained, semantically rich roles (e.g., \emph{name mover} or \emph{S-inhibition} heads) that are tightly bound to a specific node and task, and therefore do not provide a general abstraction that can be shared across tasks or embedded into an automated pipeline. (2) \textbf{Subjective functional roles.} Current interpretation methods heavily rely on subjective human judgment, making functional roles difficult to formalize and integrate into automated discovery pipelines.

To address these challenges, we propose \textbf{S}elf-\textbf{s}upervised \textbf{S}mart \textbf{Cir}cuit Discovery (S³martCirc), a framework that jointly performs circuit discovery and functional interpretation. Our approach is motivated by a key observation about LLMs: their core computation, matrix multiplication, primarily enables two fundamental operations, computational transformation and information propagation.

We therefore abstract the fine-grained, task-specific interpretations of prior work into a coarser dichotomy of \emph{computational roles} that describes \emph{how} a node processes information rather than \emph{what} content it processes. This abstraction deliberately trades semantic specificity for two properties that fine-grained roles lack: generality and quantifiability. These properties allow the abstraction to be embedded directly into an optimization objective, enabling S³martCirc to simultaneously identify important nodes and assign functional roles while explicitly modeling the interdependence between importance and functionality. Our contributions are summarized as follows:
\begin{itemize}
    \item \textbf{General functional roles:} We introduce a dichotomy of computational roles that abstracts prior task-specific interpretations into two general, quantifiable categories.
    \item \textbf{S}³\textbf{martCirc Framework:} We propose a novel MI framework that couples discovery and interpretation through bidirectional, alternating optimization.
    \item \textbf{Empirical validation:} Extensive experiments across multiple LLM architectures demonstrate the superiority of S³martCirc in identifying task-relevant circuits and recovering known circuits aligned with human-validated mechanistic interpretations.
\end{itemize}

%% file: sections/2-preliminaries.tex
In this section, we first introduce the notation used throughout the paper. We then examine circuits from two distinct tasks and define two general functional roles that characterize how nodes process information at a high level.

Given an LLM, let $P$ be its output probability distribution and $V$ be its vocabulary. We model the LLM as a computational graph in which each node is a model component (i.e. attention head or feedforward layer neuron) and each edge represents information flowing between components through the residual stream, as illustrated in Figure~\ref{fig:overview}. A circuit is the subgraph within the computational graph responsible for the model's performance on a given task. Each node takes the previous layer's representation $X \in \mathbb{R}^{T \times d_m}$ as input and produces an output $Y\in \mathbb{R}^{T \times d_n}$, where $T$ is the number of input tokens, $d_m$ is the model's hidden dimension, and $d_n$ is the node's output dimension. For attention heads, $d_n$ is determined by the architecture. For feedforward neurons, $d_n = 1$. All discovery methods we compare operate over the same node set and graph abstraction to ensure a fair comparison.

\begin{figure}
  \begin{center} \includegraphics[width=0.45\textwidth]{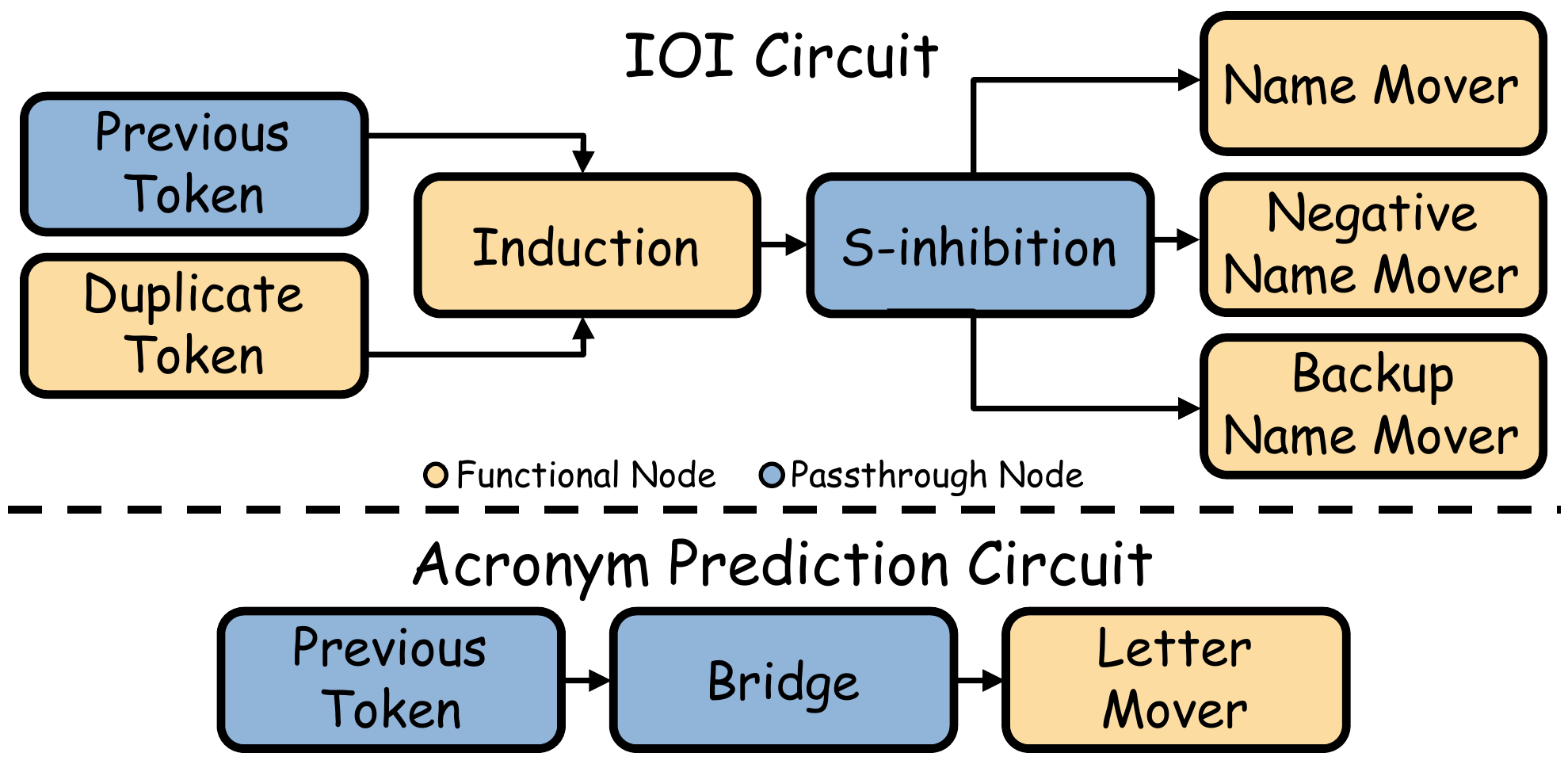}
  \end{center}
  \caption{Mapping between node-specific functional roles in discovered circuits and the defined general functional roles.}\label{fig:ioi_acro_circuit}
  \vspace{-5pt}
\end{figure}

Current MI research typically identifies and interprets nodes in the context of a single task during functional interpretation. However, when comparing circuits across different tasks, we observe striking similarities in their functional interpretations, echoing evidence that circuit components are reused across tasks with consistent roles~\cite{merullo2024circuit}. This suggests that the same small set of computational roles recurs across tasks. For example, consider the two tasks shown in Figure~\ref{fig:ioi_acro_circuit}: the Indirect Object Identification (IOI) task~\cite{wang2022interpretability}, where the model identifies the indirect object in sentences (e.g., "When John and Mary went to the store, John gave a drink to [Mary]"), and the Acronym Prediction task~\cite{garcia2024does}, where the model predicts the last letter of a three-letter acronym from a word sequence (e.g., "The Three Letter Acronym" $\rightarrow$ "TLA"). Although these tasks are distinct, their circuits share common functional interpretations, such as previous token and mover heads. 

Thus, we propose grouping node functionalities into two broad categories based on their computational role: computational transformation and information propagation. We refer to nodes in the former category as \emph{functional nodes}, which actively transform their inputs, producing outputs that differ substantially in representation or semantic content. In contrast, \emph{passthrough nodes} primarily relocate input features to downstream components with minimal modification. Importantly, this distinction captures how nodes operate on information rather than what information they process, enabling a task-agnostic characterization of node behavior.

Revisiting Figure~\ref{fig:ioi_acro_circuit}, applying this dichotomy reveals consistent patterns across both circuits. In the IOI circuit, the S-inhibition heads and previous token heads act as passthrough nodes: the former route subject-related information to downstream components, while the latter write information about the previous token into the residual stream. In contrast, the name mover heads and negative name mover heads act as functional nodes that transform this information to selectively amplify specific names in the output logit distribution. A similar decomposition arises in the acronym circuit, where previous token heads and bridge heads propagate letter information, and letter mover heads transform it to emphasize the correct output. This consistent functional decomposition across tasks with different objectives suggests systematic design principles underlying how LLMs decompose complex tasks into simpler computational components.

%% file: sections/3-method.tex
\begin{figure*}
    \centering
    \includegraphics[width=\textwidth]{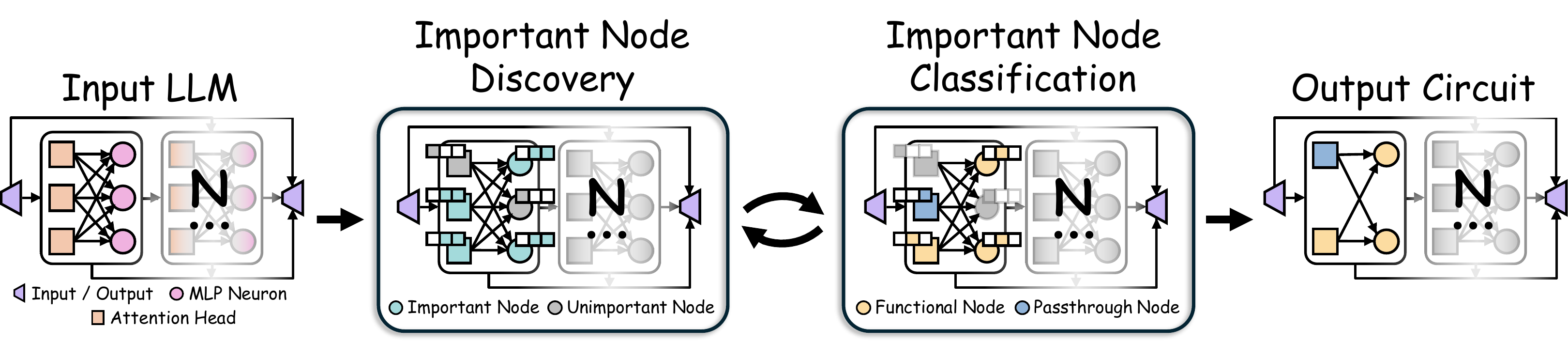}
    \caption{An overview of the proposed S³martCirc framework.}
    \label{fig:overview}
\vspace{-5pt}
\end{figure*}

In this section, we first present an overview of the proposed framework, S³martCirc. We then describe the implementation of each stage in detail and outline the training process.

\subsection{S\textmd{³}martCirc Overview}
To unify circuit discovery and functional interpretation, we jointly model both node importance and its computational role. Specifically, for every node $n$ in the LLM, we introduce a set of learnable masking coefficients $c_n = \{c_{n,u}, c_{n,p}, c_{n,f}\}$, where $c_{n,u}$ denotes the probability that node $n$ is unimportant, $c_{n,p}$ denotes the probability that the node acts as a passthrough node, and $c_{n,f}$ denotes the probability that it acts as a functional node. These coefficients are constrained such that $c_{n,u} + c_{n,p} + c_{n,f} = 1$. We optimize these coefficients through the two-stage procedure shown in Figure~\ref{fig:overview}. In the first stage, \textbf{Important Node Discovery}, we identify nodes that contribute to task performance by distinguishing nodes assigned to either functional role introduced earlier ($c_{n,p} + c_{n,f}$) from nodes classified as unimportant ($c_{n,u}$). Nodes satisfying $c_{n,p} + c_{n,f} > c_{n,u}$ are selected as candidate circuit nodes. In the second stage, \textbf{Important Node Classification}, each candidate node is assigned to one of the two roles, determined using a task-dependent metric described later. During training, node masking is applied softly using the coefficients $c_n$, while hard assignments are used only during evaluation through the following indicator function:
\begin{equation}
    \hat{c}_i =
    \begin{cases}
    1, & i = \arg\max_j c_j \\
    0, & \text{otherwise}
    \end{cases}
\end{equation}
   
\subsection{Important Node Discovery}
A circuit contains the nodes responsible for performing the task at hand. Consequently, removing any truly important node from the circuit should substantially degrade task performance. We identify important nodes by optimizing two complementary objectives: (1) minimizing the KL divergence between the circuit's output distribution and that of the full model to ensure that the circuit faithfully reproduces the model's original computation; and (2) minimizing cross-entropy loss with respect to the ground-truth answer to ensure the circuit produces correct predictions.

Formally, let $P_{\text{orig}}$ denote the output probability distribution of the full LLM, and $P_{\text{mask}}$ denote the distribution obtained when unimportant nodes are masked (i.e., nodes where $c_{n,u} > c_{n,p} + c_{n,f}$). The combined objective becomes
\begin{equation}
    L_{\text{disc.}} = \alpha \cdot \sum_{v \in V} P_{\text{orig}}(v)
    \log \frac{P_{\text{orig}}(v)}{P_{\text{mask}}(v)} - (1 - \alpha) \cdot (\log P_{\text{mask}}(y)).
\end{equation}
Here, $v\in V$ represents a token in the LLM's vocabulary, and $P_{\text{orig}}(v)$ and $P_{\text{mask}}(v)$ represent the probability assigned to token $v$ by the full model and the model when unimportant nodes are masked, respectively. $y\in V$ is the ground truth token, and $\alpha \in [0,1]$ controls the trade-off between faithfulness to the original model behavior and task performance.

\subsection{Important Node Classification}
As defined earlier, circuit nodes are categorized as functional or passthrough nodes. Prior work typically assigns these roles using qualitative analysis or heuristic criteria~\cite{wang2022interpretability, chandna2025dissecting}. In contrast, we propose a quantitative metric based on the following observation: nodes that perform nontrivial computation should alter the relational structure between tokens, whereas nodes that propagate information should preserve this structure. To formalize this idea, we measure how the pairwise similarity between tokens changes after applying a node. Intuitively, passthrough nodes induce minimal similarity changes, while functional nodes produce larger deviations.

We define a token-wise similarity operator $\text{Sim}(A)$ for an activation matrix $A \in \mathbb{R}^{T\times d_n}$ (i.e. the node's input or output) as $\text{Sim}(A) = \hat{A}\hat{A}^T \odot (\mathbf{1}\mathbf{1}^T - I).$ $\hat{A}$ denotes the row-wise $L_2$ normalization of $A$, $\mathbf{1}$ is a column vector of ones, and $I$ denotes an identity matrix. We apply element-wise multiplication with $(\mathbf{1}\mathbf{1}^T - I)$ to exclude self-similarity. Using this operator, we define the functional interpretation metric as the normalized change in token-wise similarity:
\begin{equation}
    \text{FI}(X,Y) = \frac{\|\text{Sim}(X) - \text{Sim}(Y)\|_F}{\|\text{Sim}(X)\|_F},
\end{equation}
where $X \in \mathbb{R}^{T \times d_m}$ and $Y\in \mathbb{R}^{T \times d_n}$ denote the input and output of a node, respectively. Let $N_i$ denote the set of nodes identified as important (i.e., those satisfying $c_{n,p} + c_{n,f} > c_{n,u}$). We incorporate this metric into the objective for the second stage:
\begin{align}
L_{\text{class.}} = \sum_{n_i \in N_i} \Big[
    - c_{n_i,f} \cdot \text{FI}(X_{n_i}, Y_{n_i}) 
    - \frac{c_{n_i,p} }{\text{FI}(X_{n_i}, Y_{n_i})}
\Big].
\end{align}
This objective encourages nodes that induce larger changes in similarity structure to be assigned higher $c_{n,f}$ values, while nodes that preserve similarity are assigned higher passthrough probability $c_{n,p}$. In this way, functional roles are inferred directly from measurable changes in representation, enabling a fully quantitative interpretation.

\subsection{S³martCirc Training Process}
Due to the vast size of LLMs, we introduce coefficient regularization, applied to both stages, to encourage the identification of minimal circuits while maximizing task performance. The objective consists of three terms that promote (1) confident assignments by enforcing binary values, (2) valid probability distributions over node roles (i.e., coefficients sum to one), and (3) sparse circuit selection:
\begin{equation}
\begin{aligned}
    L_{\text{sparsity}} &= \frac{1}{|C|}\sum_{c_n\in C}\Big( \underbrace{\sum_{v\in c_n}\Big[\frac{1}{2}|v(1-v)|\Big]}_{\text{confidence}} \\
    &+ \underbrace{\frac{1}{2}|1 - (c_{n,u} + c_{n,p} + c_{n,f})|}_{\text{validity}} \\
    &+ \underbrace{(c_{n,p} + c_{n,f})}_{\text{sparsity}}\Big),
\end{aligned}
\end{equation}
where $C$ is the set of all masking coefficients in the model. The first term pushes each coefficient toward binary values (0 or 1), the second term enforces that the three probabilities sum to one, and the third term directly penalizes nodes being included in the circuit, encouraging sparsity. 

As illustrated in Figure~\ref{fig:overview}, training alternates between the two stages to capture the interdependence between node importance and functional role, allowing bidirectional influence. However, assigning functional roles with randomly initialized coefficients is unreliable at the start of training. Therefore, we introduce a warmup phase that optimizes only the discovery objective ($L_{\text{disc.}}$) to first identify important nodes. After warmup, we alternate between both stages to train the masking coefficients.

%% file: sections/4-experiments.tex
\begin{table*}[t]\centering
  \caption{Performance of S³martCirc compared to baselines over three runs. Best results are shown in \textbf{bold}, and runner-up results are \textit{italicized}. Empty cells indicate that no circuit could be found within the set of nodes selected by the method. An asterisk (*) indicates that only one experimental run was completed due to the time limit (48 hours).}
  \setlength{\tabcolsep}{2pt}
  \renewcommand{\arraystretch}{1.1}
  \label{tab:main_exps}
  \resizebox{\textwidth}{!}{
  \begin{tabular}{cccccccc}
  \toprule
                         &            & \multicolumn{3}{c}{Acronym}                                               & \multicolumn{3}{c}{IOI}                                                  \\ \midrule
                         &            & Circuit Size           & Acc.                    & Logit.                 & Circuit Size          & Acc.                    & Logit.                 \\ \midrule
  \multirow{7}{*}{GPT-2}                  & Act.       & $\mathit{35.00_{\pm 4.32}}$     &  -$\mathit{54.67_{\pm 38.33}}$    & -$\mathit{3.54_{\pm 38.33}}$    & $39.33_{\pm 10.40}$         & -$\mathit{70.71_{\pm 25.89}}$    & -$\mathit{9.11_{\pm 25.89}}$   \\
                         & Attr.      & $37.33_{\pm 2.05}$           & -$0.17_{\pm 0.24}$            & $0.46_{\pm 0.24}$            & $48.00_{\pm 2.16}$          & -$12.73_{\pm 8.01}$           & -$5.14_{\pm 8.01}$           \\
                         & Attr. IG   & $49.67_{\pm 11.44}$          & -$2.00_{\pm 1.08}$            & -$0.59_{\pm 1.08}$           & $\mathbf{38.00_{\pm 5.10}}$ & -$0.17_{\pm 0.24}$            & -$0.29_{\pm 0.24}$           \\
                         & IBCircuit  & \---                   & \---                    & \---                   & \---                  & \---                    & \---                   \\
                         & Prune.     & $\mathbf{32.33_{\pm 37.97}}$ & $0.00_{\pm 0.00}$             & $0.07_{\pm 0.00}$            & $\mathit{38.33_{\pm 33.81}}$   & $0.00_{\pm 0.00}$             & -$0.02_{\pm 0.00}$           \\
                         & Random     & $60.33_{\pm 7.76}$           & $0.00_{\pm 0.00}$             & $0.04_{\pm 0.00}$            & $77.33_{\pm 3.40}$          & $0.00_{\pm 0.00}$             & -$0.11_{\pm 0.00}$           \\
                         & S³martCirc & $60.33_{\pm 7.93}$           & $\mathbf{-97.50_{\pm 1.08}}$  & $\mathbf{-5.00_{\pm 1.08}}$  & $76.00_{\pm 5.72}$          & $\mathbf{-95.99_{\pm 2.85}}$  & $\mathbf{-10.42_{\pm 2.85}}$ \\ \midrule
  \multirow{7}{*}{Llama} & Act.       & \---                   & \---                    & \---                   & $\mathit{13.00_{\pm 2.00}}$    & -$\mathit{59.17_{\pm 41.90}}$    & -$\mathit{2.37_{\pm 41.90}}$    \\
                         & Attr.      & \---                   & \---                    & \---                   & $71.67_{\pm 2.05}$          & -$1.50_{\pm 0.41}$            & -$0.35_{\pm 0.41}$           \\
                         & Attr. IG   & $54.67_{\pm 2.62}$           & -$\mathit{2.17_{\pm 0.24}}$      & -$\mathit{0.34_{\pm 0.24}}$     & $47.67_{\pm 3.30}$          & -$5.67_{\pm 3.70}$            & -$0.35_{\pm 3.70}$           \\
                         & IBCircuit  & $\mathbf{29.67_{\pm 12.66}}$ & $0.00_{\pm 0.00}$             & -$0.08_{\pm 0.00}$           & $\mathbf{7.33_{\pm 0.94}}$  & $0.00_{\pm 0.00}$             & $0.00_{\pm 0.00}$            \\
                         & Prune.     & $70.67_{\pm 3.09}$           & $0.00_{\pm 0.00}$             & $0.01_{\pm 0.00}$            & $70.67_{\pm 3.09}$          & $0.00_{\pm 0.00}$             & $0.01_{\pm 0.00}$            \\
                         & Random     & $60.00_{\pm 9.93}$           & $0.00_{\pm 0.00}$             & -$0.00_{\pm 0.00}$           & $62.33_{\pm 6.18}$          & $0.00_{\pm 0.00}$             & -$0.01_{\pm 0.00}$           \\
                         & S³martCirc & $\mathit{47.33_{\pm 2.05}}$     & $\mathbf{-57.33_{\pm 3.40}}$  & $\mathbf{-3.17_{\pm 3.40}}$  & $35.00_{\pm 3.74}$          & $\mathbf{-69.00_{\pm 42.43}}$ & $\mathbf{-2.96_{\pm 42.43}}$ \\ \midrule
  \multirow{7}{*}{Qwen}  & Act.       & $32.00_{\pm 0.00}$\textsuperscript{*}           & $\mathbf{-100.00_{\pm 0.00}}$\textsuperscript{*} & -$\mathit{1.00_{\pm 0.00}}$\textsuperscript{*}     & $\mathbf{12.00_{\pm 0.00}}$\textsuperscript{*} &  -$\mathit{33.00_{\pm 0.00}}$\textsuperscript{*}     & -$0.33_{\pm 0.00}$\textsuperscript{*}           \\
                         & Attr.      & $64.33_{\pm 3.30}$           & -$\mathit{0.17_{\pm 0.24}}$      & -$0.48_{\pm 0.24}$           & $29.00_{\pm 3.56}$          & -$0.50_{\pm 0.41}$            & -$\mathit{0.91_{\pm 0.41}}$     \\
                         & Attr. IG   & $32.33_{\pm 8.34}$           & -$\mathit{0.17_{\pm 0.24}}$      & -$0.12_{\pm 0.24}$           & $35.00_{\pm 5.35}$          & -$0.33_{\pm 0.47}$            & -$0.19_{\pm 0.47}$           \\
                         & IBCircuit  & $\mathit{14.67_{\pm 0.94}}$     & $0.00_{\pm 0.00}$             & $0.29_{\pm 0.00}$            & \---                  & \---                    & \---                   \\
                         & Prune.     & $46.00_{\pm 4.24}$           & $0.00_{\pm 0.00}$             & -$0.01_{\pm 0.00}$           & $46.00_{\pm 4.24}$          & $0.00_{\pm 0.00}$             & -$0.01_{\pm 0.00}$           \\
                         & Random     & $52.67_{\pm 9.74}$           & $0.00_{\pm 0.00}$             & -$0.00_{\pm 0.00}$           & $48.00_{\pm 6.98}$          & $0.00_{\pm 0.00}$             & $0.01_{\pm 0.00}$            \\
                         & S³martCirc & $\mathbf{12.00_{\pm 1.63}}$  & $\mathbf{-100.00_{\pm 0.00}}$ & $\mathbf{-10.97_{\pm 0.00}}$ & $\mathit{16.33_{\pm 4.50}}$    & $\mathbf{-100.00_{\pm 0.00}}$ & $\mathbf{-12.01_{\pm 0.00}}$ \\ \bottomrule
  \end{tabular}
  }
\vspace{-10pt}
\end{table*}
  
In this section, we first introduce the experiment setup. Then, we answer the following research questions: \textbf{RQ1.} How effective is S³martCirc at identifying circuits compared to existing methods? \textbf{RQ2.} Is S³martCirc able to recover known circuits from prior work? \textbf{RQ3.} How sensitive is S³martCirc to varying hyperparameters? \textbf{RQ4.} How does each component contribute to the performance of S³martCirc? 

\subsection{Experiment Setup}
\noindent\textbf{LLMs.} We select three LLMs of varying sizes for our experiments: GPT-2~\cite{radford2019language, hf_canonical_model_maintainers_2022}, Llama 3.2 1B~\cite{grattafiori2024llama, llama321b}, and Qwen 3 1.7B~\cite{qwen3technicalreport, qwen317b}. GPT-2 has been extensively examined in prior MI work, making it a well-understood and widely used benchmark for circuit analysis. Llama 3.2 1B and Qwen 3 1.7B are included to evaluate whether S³martCirc generalizes across model architectures.

\noindent\textbf{Evaluation Tasks.} We collect two tasks to perform circuit discovery on: Indirect Object Identification (IOI)~\cite{wang2022interpretability} and Acronym Prediction (Acronym)~\cite{garcia2024does}. Both tasks are drawn from previous MI work where extensive experiments have been performed to uncover the task circuit in GPT-2. 

\noindent\textbf{Evaluation Metrics.} We use three metrics: \textbf{Circuit Size}, \textbf{Accuracy Drop (Acc.)}, and \textbf{Logit Difference (Logit.)}. Each method selects a set of important nodes. We then remove nodes that are not connected to the input or output, yielding the final end-to-end circuit; \textbf{Circuit Size} is defined as the size of this circuit. Unlike during training, we evaluate circuit performance by masking only the circuit nodes in the model, denoted as $P_{\text{circ}}$. \textbf{Acc.} measures the percent drop in accuracy between the original model and $P_{\text{circ}}$. Lastly, \textbf{Logit.} denotes the difference in the logit assigned to the original model’s predicted answer between the original model and $P_{\text{circ}}$.

\noindent\textbf{Baselines.}
We adopt six circuit discovery methods as baselines. We first consider patching-based approaches: \textbf{Activation Patching (Act.)}~\cite{heimersheim2024use} measures each node’s contribution by ablating it (i.e., replacing its activation with a corrupted counterpart) and evaluating the resulting change in logit difference between the model’s original answer and an alternative response. \textbf{Attribution Patching (Attr.)}~\cite{syed2023attribution} provides a linear approximation of each node’s contribution by multiplying the difference between its clean and corrupted activations with the gradient of a logit-based metric (i.e., logit difference between the original and corrupted outputs). \textbf{Attribution Patching with Integrated Gradients (Attr. IG)}~\cite{hanna2024have} extends attribution patching by integrating gradients along a path between the corrupted and clean activations, yielding a more accurate attribution score. We next consider optimization-based approaches: \textbf{IBCircuit}~\cite{bian2026ibcircuit} performs circuit discovery by optimizing an information bottleneck objective to identify compact, task-relevant subgraphs. \textbf{Pruning (Prune.)}~\cite{bhaskar2024finding} applies gradient-based pruning to remove less important edges; we adapt this method to operate at the node level for fair comparison. \textbf{Random} serves as a lower bound by randomly selecting $k$ nodes as the circuit.

\subsection{RQ1: Performance of S³martCirc}
In Table~\ref{tab:main_exps}, we compare S³martCirc to current circuit discovery methods and observe the following: 

\noindent{\textbf{Effectiveness.}} S³martCirc outperforms all baselines by substantial margins across all tasks and models. For example, on Acronym with GPT-2, S³martCirc achieves an accuracy drop of -97.50\%, compared to -54.67\% for the strongest baseline, Activation Patching. These results indicate that an optimizable mask is more effective in identifying nodes critical to model computation than patching-based methods, which evaluate nodes independently and potentially miss important interactions between components. Furthermore, although Prune. and IBCircuit also employ continuous masking strategies, both methods perform substantially worse than S³martCirc. This suggests that incorporating functional interpretation provides essential guidance for identifying circuit nodes, beyond what is achievable through masking alone.

\noindent{\textbf{Circuit Size.}} Our goal is to identify a minimal circuit that achieves maximal accuracy drop and logit difference. While S³martCirc does not consistently produce the smallest circuits, it achieves substantially greater performance degradation for comparable circuit sizes, indicating more effective node selection than baseline methods. For example, on the IOI task with Llama, S³martCirc achieves a -69.00\% accuracy drop using a circuit of 35 nodes. In contrast, Attr. IG produces a larger circuit (average size 47.67 nodes) but yields only a -5.67\% accuracy drop. This shows that S³martCirc identifies more functionally critical nodes and, again, implies the benefit of incorporating functional interpretation.

\subsection{RQ2: Recovering Prior Circuits}
\begin{figure}
  \begin{center}
    \includegraphics[width=0.4\textwidth]{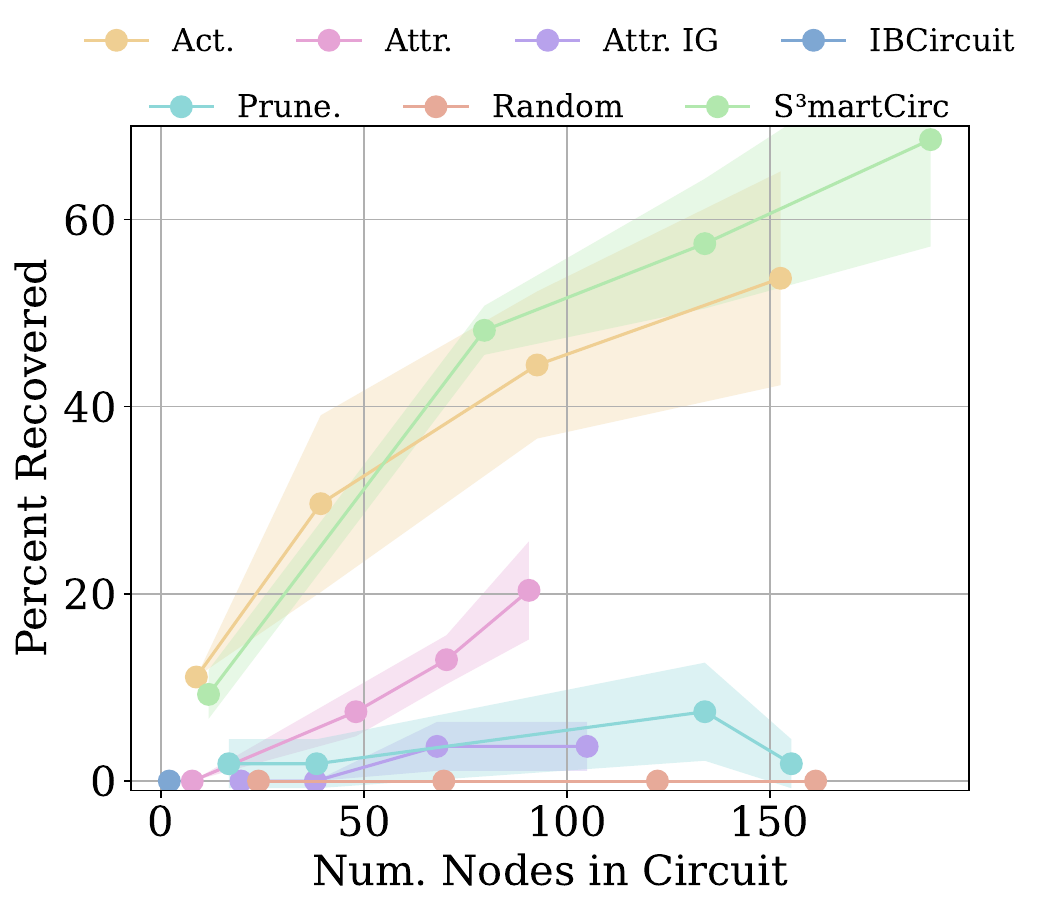}
  \end{center}
  \caption{Percent of the IOI circuit recovered by different circuit discovery methods}\label{fig:ioi_comparison}
  \vspace{-5pt}
\end{figure}

Previous work~\cite{wang2022interpretability} manually identified and verified a ground-truth circuit for the IOI task through extensive analysis and validation. This circuit serves as a reliable reference for evaluating whether automated methods can recover the same critical components. Accordingly, a desirable property of circuit discovery methods is the ability to recover nodes from this known IOI circuit. Figure~\ref{fig:ioi_comparison} reports the percentage of nodes from the original IOI circuit that are recovered by each method within the final end-to-end discovered circuits. We evaluate this across varying circuit sizes obtained by thresholding at 50, 100, 150, and 200 nodes. Note that perfect recovery (100\%) is not achievable due to the vast number of possible nodes to choose from.

Overall, S³martCirc consistently recovers a higher percent of the original circuit compared to most baselines, and its performance is comparable only to Activation Patching. In contrast, other baselines, particularly optimization-based methods, achieve substantially lower recovery rates, even when allowed to select up to 200 nodes. These results provide evidence that S³martCirc not only improves circuit discovery performance (\textbf{RQ1}), but also aligns more closely with human-verified mechanistic interpretations of the IOI task. This increases the credibility of its discovered circuits, particularly in settings where ground-truth labels are unavailable.

\begin{figure}
  \begin{center}
    \includegraphics[width=0.3\textwidth]{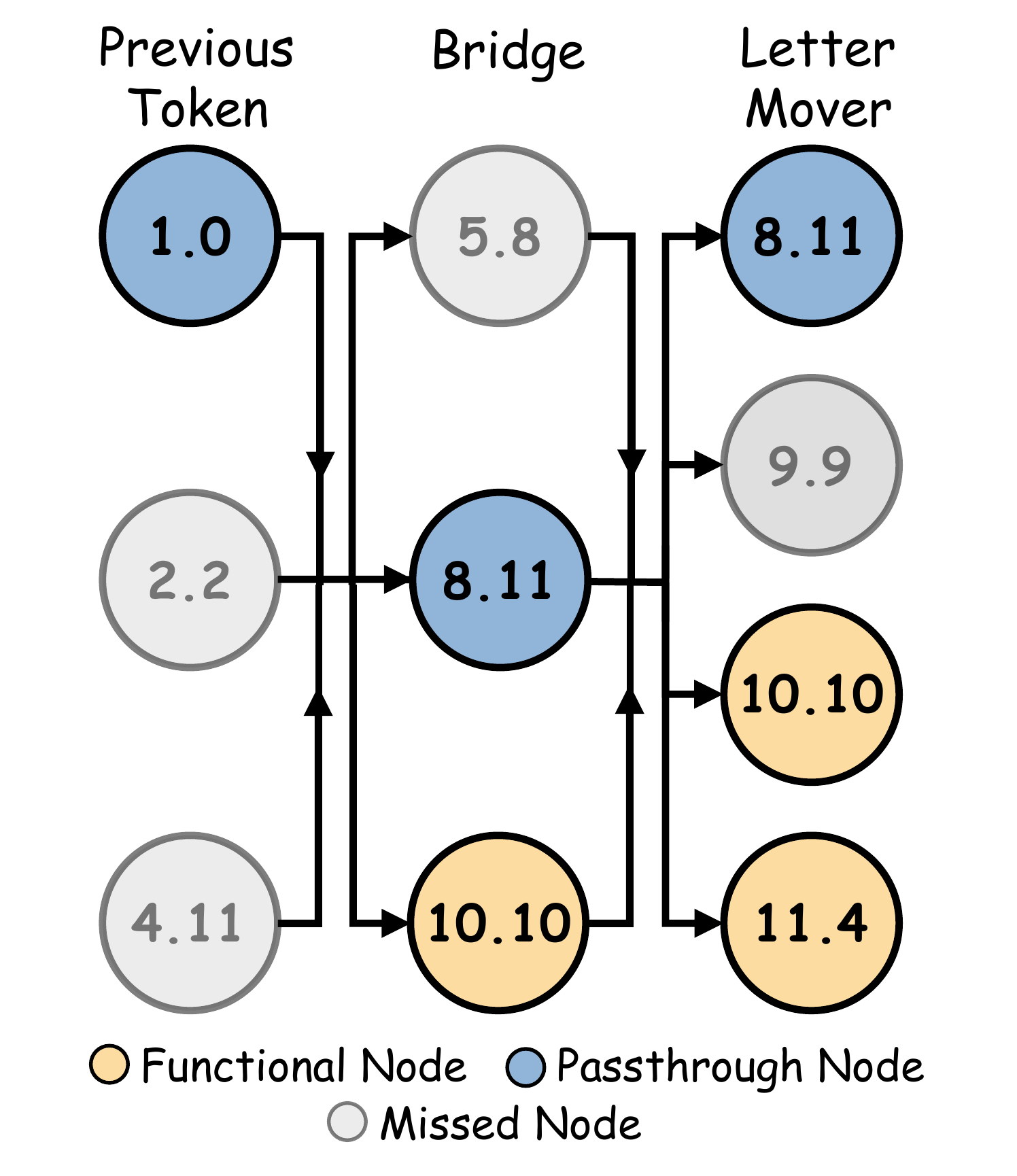}
  \end{center}
  \caption{Comparison between the Acronym circuit and the functional interpretations determined by S³martCirc.}\label{fig:acro_fi}
  \vspace{-5pt}
\end{figure}

We further validate the performance of S³martCirc by comparing its functional interpretations against the ground-truth circuit for the Acronym Prediction task~\cite{garcia2024does}, shown in Figure~\ref{fig:acro_fi}. In the figure, each node is labeled as layer.head (e.g., 8.11 represents layer 8, attention head 11). S³martCirc successfully recovers four of the eight nodes (50\%) from the original circuit, and its functional classifications align with the mapping shown in Figure~\ref{fig:ioi_acro_circuit}. Specifically, Previous Token node 1.0 is correctly identified as a passthrough node, while Letter Mover node 11.4 is correctly classified as a functional node. This alignment, again, indicates that S³martCirc not only identifies important nodes but also accurately captures their mechanistic roles, providing automated interpretations that correspond closely to those derived through human analysis. 

However, an apparent discrepancy arises in the classification of nodes 8.11 and 10.10. S³martCirc classifies 8.11 as a passthrough node while classifying 10.10 as a functional node, even though prior work shows that both heads can act as either Bridge nodes or Letter Mover nodes depending on the specific intervention applied. This divergence highlights a fundamental difference in methodology: human-verified approaches evaluate node behavior under carefully isolated interventions to determine what functions a head can perform, while S³martCirc measures the dominant contribution of each head across multiple samples. In other words, S³martCirc assigns labels based on the function that most strongly influences the circuit’s overall behavior. The differing classifications therefore reflect the multi-functional nature of these heads rather than a misclassification.

\subsection{RQ3: Parameter Analysis}
\begin{figure}
  \centering
  \includegraphics[width=0.9\columnwidth]{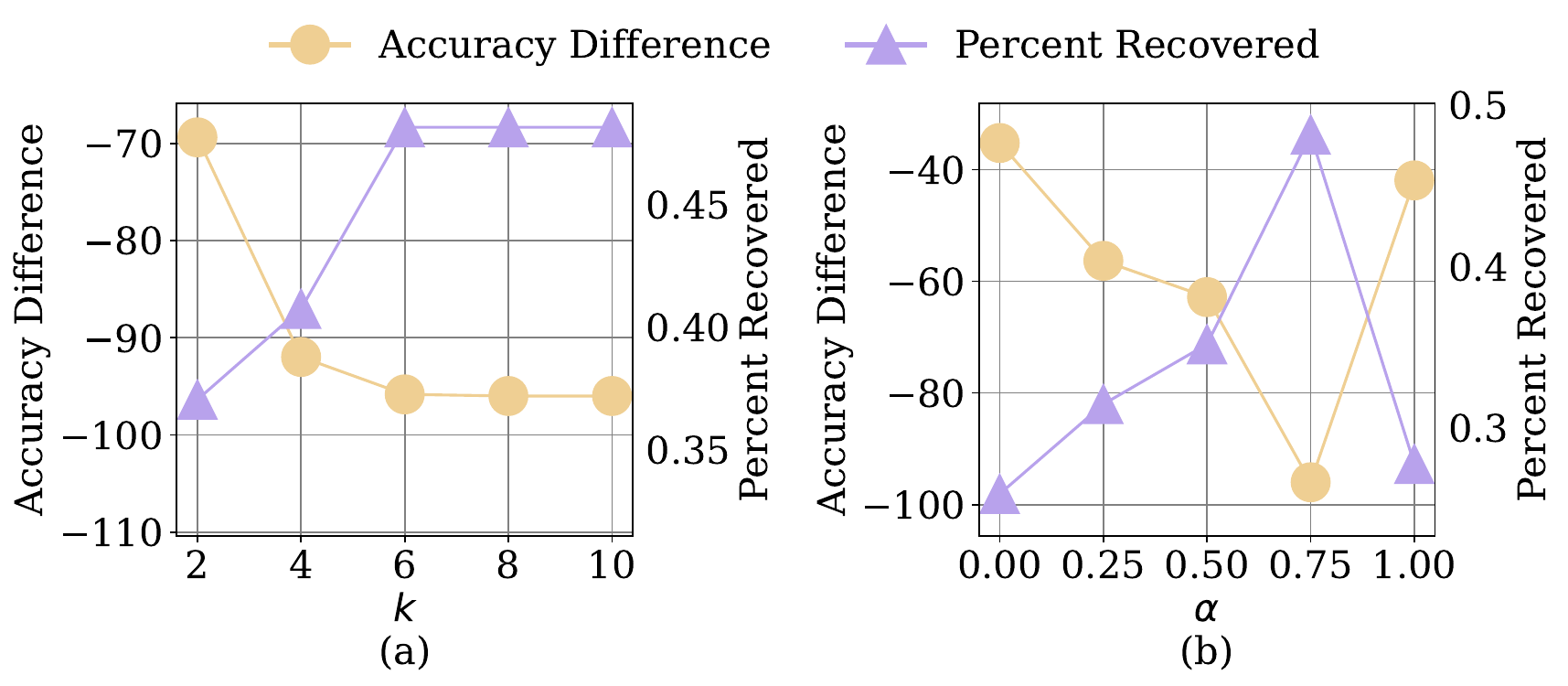}
  \caption{Effect of $\alpha$ and $k$ on S³martCirc Performance.}
  \label{fig:param_analysis}
\end{figure}

To answer \textbf{RQ3}, we perform a parameter analysis, shown in Figure~\ref{fig:param_analysis}, on the IOI task using GPT-2 to evaluate the sensitivity of S³martCirc to two key hyperparameters: $\alpha$, which balances the two objectives in the Important Node Discovery stage, and $k$, which controls the number of steps in the Important Node Classification stage. Specifically, each iteration consists of one discovery step followed by $k$ classification steps. We evaluate both accuracy (Acc.) and the percentage of recovered nodes from the original IOI circuit across $k \in \{2, 4, 6, 8, 10\}$ and $\alpha \in \{0.0, 0.25, 0.5, 0.75, 1.0\}$.

Figure~\ref{fig:param_analysis}a demonstrates that performance generally improves as $k$ increases, and stabilizes at $k=6$. The accuracy drop decreases from -71\% at $k=2$ to approximately -96\% at $k=6$, and percent circuit recovery monotonically increases. The consistent improvement suggests that additional refinement steps in the classification stage improve the identification of important nodes and the circuit quality.

Figure~\ref{fig:param_analysis}b reveals that $\alpha = 0.75$ achieves the best performance in terms of both accuracy and circuit recovery. This indicates that, for the IOI task, the KL divergence objective should be weighted more heavily than the task loss during circuit discovery. Nevertheless, the cross-entropy term remains important, as performance degrades sharply when it is removed ($\alpha = 1.0$). This helps explain the weaker performance of baselines, which rely primarily on output distribution differences (e.g., KL divergence) to identify important nodes. Our results suggest that, for IOI, combining output faithfulness with task performance signals is more effective, as some critical components contribute to correct behavior without inducing large changes in logits.

\subsection{RQ4: Ablation Study}
\begin{figure}
  \begin{center}
    \includegraphics[width=0.4\textwidth]{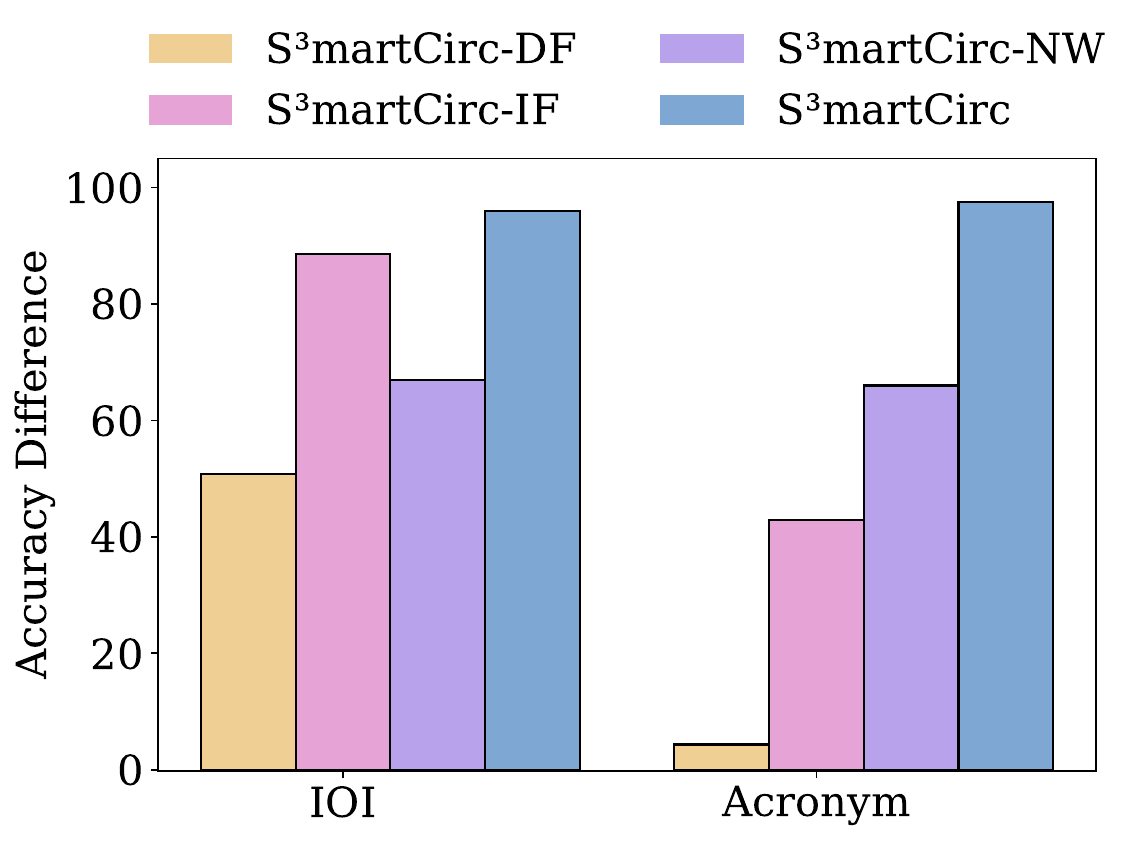}
  \end{center}
  \caption{Ablation study to evaluate the effectiveness of different components in S³martCirc and its training process.}\label{fig:ablation}
  \vspace{-5pt}
\end{figure}

We perform an ablation study to understand how each component of S³martCirc contributes to its overall performance. Figure~\ref{fig:ablation} compares the accuracy difference achieved by the original S³martCirc method against three variants: \textbf{S}³\textbf{martCirc-NW} (\textbf{N}o \textbf{W}armup), which removes the initial warmup period and directly starts the alternating training process; \textbf{S}³\textbf{martCirc-DF} (\textbf{D}iscovery \textbf{F}irst), which completes the entire Important Node Discovery stage before beginning Important Node Classification; and \textbf{S}³\textbf{martCirc-IF} (\textbf{I}nterpretation \textbf{F}irst), which reverses the order by performing Important Node Classification before Important Node Discovery. These ablations isolate the contributions of the warmup period and the alternating training strategy.

\noindent\textbf{Importance of Alternating Training.} 
S³martCirc-DF performs substantially worse than the other variants, and achieves near-zero accuracy drop on Acronym (-4.33\% compared to -97.5\% for S³martCirc). This sharp degradation suggests that identifying all important nodes prior to functional interpretation leads to an overly broad candidate set, making it difficult to distinguish truly critical nodes from less relevant ones. As a result, applying functional interpretation only after discovery becomes less effective, as it is overwhelmed by noise from the large set of candidate nodes. This leads to circuits that include non-critical components while omitting key ones.

S³martCirc-IF performs significantly better than S³martCirc-DF (approximately 88\% on IOI and 42\% on Acronym), indicating that incorporating functional information into the discovery process provides useful guidance for identifying important nodes. However, it still underperforms S³martCirc. This demonstrates that, while functional interpretation is beneficial as a prior, it is insufficient without the alternating training. 

Overall, these results highlight the importance of bidirectional coupling between Important Node Discovery and Important Node Classification. The alternating optimization allows each stage to refine the other: discovery benefits from evolving functional understanding, while classification operates on progressively more focused candidate sets.

\noindent\textbf{Role of Warmup Period.} The removal of the warmup period (S³martCirc-NW) affects performance consistently across tasks (67\% on IOI and 66\% on Acronym). This suggests that directly starting alternating optimization from randomly initialized masking coefficients provides an unreliable signal for early-stage updates. In particular, without an initial phase that identifies a reasonable set of candidate nodes, the model struggles to generate informative gradients for distinguishing important from unimportant components. As a result, the optimization process is less stable and converges to suboptimal circuits, making it more difficult to recover the underlying task-relevant structure.

%% file: sections/5-related_works.tex
\noindent\textbf{Mechanistic Interpretability and Circuit Discovery.}
Mechanistic interpretability (MI) explains model behavior by identifying the components responsible for it~\cite{bereska2024mechanistic,sharkey2025open,rai2024practical}, with the \emph{circuit} as its central abstraction. Most work follows a \emph{where-then-what} paradigm~\cite{meloux2025everything,garcia2024detecting,chandna2025dissecting}, localizing components via patching-based interventions~\cite{wang2022interpretability,meng2022locating,garcia2024does,heimersheim2024use} or gradient-based approximations~\cite{syed2023attribution,hanna2024have} and then interpreting their roles manually. Such interventions are costly and evaluate components in isolation, while the interpretation step is subjective and hard to scale. Recent work further shows that functional roles are not task-specific: components are reused across tasks with consistent roles~\cite{merullo2024circuit}, roles can be localized to individual neurons~\cite{nikankin2026same}, and structurally distinct circuits can be functionally equivalent~\cite{haklay2026pitfalls}. This motivates the abstract, quantitative notion of role that S³martCirc folds directly into discovery.

\noindent\textbf{Optimization-Based Circuit Discovery.}
A complementary line optimizes continuous masks over components under sparsity constraints~\cite{franklelottery,bhaskar2024finding}, e.g., IBCircuit's information bottleneck objective~\cite{bian2026ibcircuit} and behavior-specific mask learning~\cite{li2024optimal}. These better capture component interactions than patching, but most model only \emph{which} components matter, not \emph{how} they contribute. For example, IBCircuit formulates circuit extraction as an optimization problem balancing behavioral fidelity with an information bottleneck objective~\cite{bian2026ibcircuit}, while other approaches optimize masks to isolate behavior-specific components~\cite{li2024optimal}. S³martCirc differs in that (1) roles come from a task-agnostic, quantitative metric on activations rather than hand-specified desiderata or ablation signs; (2) importance and role are optimized jointly by alternating between stages, not read off a single mask; and (3) discovery spans both attention heads and MLP neurons.

%% file: sections/6-conclusion.tex
We introduce S³martCirc, a unified mechanistic interpretability framework that jointly discovers circuits and interprets node functionality in LLMs. We propose a generalized dichotomy of abstract functional roles that applies across tasks, together with a quantitative metric that integrates these interpretations into an end-to-end optimizable circuit discovery process. This enables S³martCirc to directly model the interdependence between node importance and functional role. Extensive experiments on GPT-2, Llama 3.2, and Qwen 3 demonstrate that S³martCirc substantially outperforms existing methods in identifying compact, task-relevant circuits and recovering human-verified circuits, advancing automated and scalable MI. Our framework also has limitations that point to future work: the two roles are intentionally coarse and do not recover the fine-grained roles of manual analysis, and because roles are defined relative to a task's activations rather than intrinsically, the same node may be labeled differently across tasks.

%% file: sections/appendix.tex
\section{Implementation Details}
\subsection{Computation Resources}
All experiments were performed on clusters of NVIDIA RTX A4000 16GB and NVIDIA A16 16GB GPUs except for Attribution Patching and Attribution Patching IG on Qwen. Due to the large model size, these experiments were done on clusters of NVIDIA RTX A6000 48GB GPUs.

\subsection{Data Generation}
We generate the data samples for each experiment due to the need to maintain token limits based on each model's tokenizer. Below describes the data generation process for each task. given a tokenizer. 

\noindent{\textbf{IOI.}} We adopt the names and place/object pairs from~\cite{wang2022interpretability}. For each model, we keep only the names and place/object pairs that tokenize as a single token, yielding a valid set of values. We then enumerate all valid combinations and insert each into the template ``When [NAME\_1] and [NAME\_2] went to the [PLACE], [NAME\_1] handed the [OBJECT] to''. Corrupted counterparts are formed by swapping the second name, producing clean/corrupted pairs. Finally, to obtain a balanced set of 1000 samples, we draw $1000/p$ samples from each place/object pair, where $p$ is the number of valid pairs.

\noindent{\textbf{Acronym.}} We generate samples using the \texttt{RandomWord} class from the \texttt{wonderwords} package. For each letter of the alphabet, we sample 500 words and keep a word only if its formatted version (i.e., prepending a space and capitalizing the first letter) tokenizes into exactly two tokens, the first of which is the space-prefixed capital letter. Letters with no valid words are discarded, leaving a set of valid letters. Within this set, we identify valid abbreviations, i.e., three-letter sequences whose concatenation tokenizes into three tokens. For each pair of valid abbreviations, we sample random words matching their letters and insert them into the template ``The [A\_RAND\_WORD] [B\_RAND\_WORD] [C\_RAND\_WORD] (AB'' to form clean and corrupted samples. Finally, we shuffle the generated data and take the first 1000 samples.

\subsection{Hyperparameters}
Below, we list the hyperparameters used for each model dataset combination in Table~\ref{tab:main_exps}. All experiments used a random seed of 42. 
\begin{itemize}
    \item Acronym with GPT-2 - Learning rate is 0.01, weight decay is 0.005, number of epochs is 60, number of warmup epochs is 60, sparsity coefficient is 0.5, and the logit coefficient (alpha) is 0.25.
    \item IOI with GPT-2  - Learning rate is 0.005, weight decay is 0.0005, number of epochs is 60, number of warmup epochs is 5, sparsity coefficient is 0.3, and the logit coefficient is 0.75.
    \item Acronym with Llama - Learning rate is 0.001, weight decay is 0.005, number of epochs is 50, number of warmup epochs is 5, sparsity coefficient is 1.0, and the logit coefficient is 0.75.
    \item IOI with Llama - Learning rate is 0.001, weight decay is 0.005, number of epochs is 75, number of warmup epochs is 5, sparsity coefficient is 1.0, and the logit coefficient is 0.25.
    \item Acronym with Qwen - Learning rate is 0.001, weight decay is 0.005, number of epochs is 50, number of warmup epochs is 5, sparsity coefficient is 0.7, and the logit coefficient is 0.75.
    \item IOI with Qwen - Learning rate is 0.001, weight decay is 0.005, number of epochs is 50, number of warmup epochs is 10, sparsity coefficient is 0.9, and the logit coefficient is 0.75.
\end{itemize}

\subsection{Run time}
In Table~\ref{tab:runtime}, we compare the runtime of S³martCirc to the baselines. From the table, we see that S³martCirc is relatively efficient, achieving high performance without requiring an excessive amount of time. 
\begin{table}[]
\caption{Runtime comparison between baselines and S³martCirc (in seconds)}\label{tab:runtime}
\resizebox{0.5\textwidth}{!}{
\begin{tabular}{ccccccccc}
\hline
                       &         & Act.      & Attr.   & Attr. IG & IBCircuit & Prune.   & Random & S³martCirc \\ \hline
\multirow{2}{*}{GPT-2} & Acronym & 6239.72   & 299.54  & 3452.48  & 426.70    & 3391.87  & 3.98   & 197.14     \\
                       & IOI     & 7943.71   & 267.94  & 2801.64  & 467.19    & 3732.88  & 0.13   & 264.44     \\ \hline
\multirow{2}{*}{Llama} & Acronym & 61510.09  & 1114.46 & 12770.96 & 1266.76   & 11470.94 & 2.22   & 507.67     \\
                       & IOI     & 64639.54  & 1128.62 & 12344.18 & 1617.78   & 11645.51 & 3.37   & 901.02     \\ \hline
\multirow{2}{*}{Qwen}  & Acronym & 168689.44 & 829.43  & 11151.09 & 3845.65   & 19072.51 & 3.54   & 982.16     \\
                       & IOI     & 187663.88 & 783.17  & 11148.12 & 5152.79   & 21000.88 & 2.36   & 1047.59    \\ \hline
\end{tabular}}
\end{table}

\section{Additional Experiment Results}
\begin{figure}
  \begin{center}
    \includegraphics[width=0.4\textwidth]{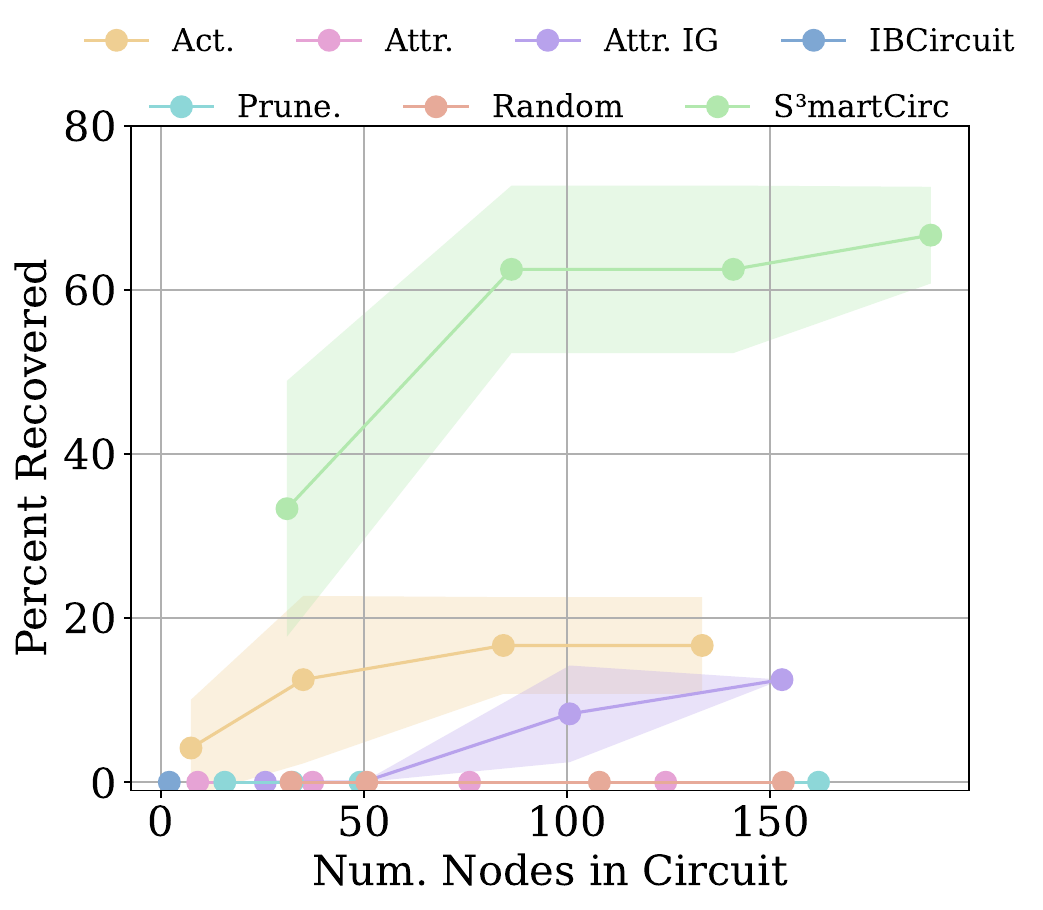}
  \end{center}
  \caption{Percentage of the Acronym circuit recovered by different circuit discovery methods}\label{fig:acro_comparison}
  \vspace{-10pt}
\end{figure}

We also repeated the first experiment in Section 4.3 for the acronym task and show the results in Figure~\ref{fig:acro_comparison}. We see that our claims still hold; S³martCirc maintains the largest percent recovered compared to the baselines. 

